\documentclass[a4paper, 10pt, conference]{ieeeconf}      
\usepackage{FG2025}
\usepackage{graphicx} 
\usepackage{xcolor}
\usepackage{gensymb}
\usepackage{algorithm}
\usepackage{algpseudocode}
\FGfinalcopy 

\IEEEoverridecommandlockouts                              
\def\FGPaperID{****} 

\title{\LARGE \bf
Dissecting Human Body Representations in 
Deep Networks Trained for Person Identification 
}

\author{\parbox{16cm}{\centering
    {\large Thomas M Metz, Matthew Q Hill, Blake Myers, Veda Nandan Gandi, Rahul Chilakapati, Alice J O’Toole}\\ 
    {\normalsize School of Behavioral and Brain Science, The University of Texas at Dallas, Richardson, Texas\\}}
    \thanks{This research is based upon work supported in part by
the Office of the Director of National Intelligence (ODNI),
Intelligence Advanced Research Projects Activity (IARPA),
via [2022-21102100005]. The views and conclusions contained herein are those of the authors and should not be
interpreted as necessarily representing the official policies,
either expressed or implied, of ODNI, IARPA, or the U.S.
Government. The US. Government is authorized to reproduce
and distribute reprints for governmental purposes notwithstanding any copyright annotation therein.}
}

\begin{document}

\ifFGfinal
\thispagestyle{empty}
\pagestyle{empty}
\else
\author{Anonymous FG2025 submission\\ Paper ID \FGPaperID \\}
\pagestyle{plain}
\fi
\maketitle

\begin{abstract}
Long-term body identification algorithms have emerged recently with the increased availability of high-quality training data. We seek to fill knowledge gaps about these models by analyzing body image embeddings from four body identification networks trained with 1.9 million images across 4,788 identities and 9 databases. By analyzing a diverse range of architectures (ViT, SWIN-ViT, CNN, and linguistically primed CNN), we first show that
the face contributes to the accuracy of body identification algorithms and that these algorithms can identify faces to some extent---with no explicit face training. Second, we show that  representations (embeddings) generated by 
body identification algorithms encode information about  gender, 
as well as image-based information including view (yaw) and 
even the dataset from which the image originated. Third, we demonstrate that identification accuracy can be improved without additional training by operating directly and selectively on the learned embedding space. Leveraging principal component analysis (PCA), identity comparisons were consistently more accurate in subspaces that eliminated dimensions that explained large amounts of variance.  These three findings were surprisingly consistent
across architectures and test datasets. This work
represents the first analysis of body representations
produced by long-term re-identification networks trained
on challenging unconstrained datasets.

\end{abstract}

\section{INTRODUCTION}


In recent years, body identification algorithms that go
beyond short-term re-identification  have been developed 
for a variety of important applications (cf. for a review \cite{ye2021deep}). These algorithms are useful when the face is not visible or
is not of sufficient quality to be identified.
A strong focus in long-term re-identifcation has been on the problem
of clothing change, (e.g., \cite{chen2021learning, gu2022clothes, hong2021fine, huang2023wholebodydetectionrecognitionidentification, liu2023learning, Liu_2024_CVPR,myers2023recognizing,yang2019person, ye2021deep}). Increasingly, body identification models have begun to tackle long-term re-identification in
unconstrained viewing 
environments (e.g.,
\cite{huang2023wholebodydetectionrecognitionidentification,Liu_2024_CVPR,myers2023recognizing}).
These models build on the expanding availability of datasets
suitable for training
body identification networks over a wide range of views and distances \cite{cornett2023expanding}. 
The body representations that result from such training
are robust to changes in clothing, viewpoint, illumination, and distance. 

Despite recent advances in body-based identification, work aimed at understanding body representations in identity-trained deep networks is surprisingly limited. 
Our goal is to provide an in-depth analysis of the information encoded in  representations that 
support long-term body identification.
 We consider three questions.
First, how much does the face contribute to the performance of
long-term body identification networks? 
This question has implications for the
security and privacy of body identification models \cite{dietlmeier2021important}.

Second,  do body recognition networks retain information
about person attributes (gender) and about imaging conditions (camera angle, dataset of origin), in addition to identity?  
We will show that they do.
The nature of information
retained in body embeddings
has implications for determining 
potential use cases for these models, including the development of
semantic editing applications 
\cite{shen2020interpreting}. 
Third, 
can editing of the latent space 
improve identification performance without additional training?

We focus on the analysis of body representations in four identity-trained body networks with diverse backbone architectures (ResNet \cite{he2015deepresiduallearningimage}, 
Vision Transformer \cite{dosovitskiy2021imageworth16x16words}
and Swin Vision Transformer \cite{liu2021swintransformerhierarchicalvision}). Different architectures were used  to distinguish between
coding principles common across  networks and those that apply to specific architectures. 
We equated training across models to ensure that
differences among model-generated representations
would not be due 
to the training data. To achieve
robust models with the most generalizable applicability, 
we compiled a large and diverse training dataset comprised of 
over 1.9 million images of nearly 5K identities across 9 datasets.

\subsection{Contributions}
\begin{itemize}
  \item We show that identification accuracy declines when the head/face region is obscured in a whole-person image, indicating that the face contributes to body-identification. 
  \item We show that networks trained for body identification are somewhat capable of face identification.  
  \item Analogous to face identification networks, we show that body embeddings retain accurate information about a person's gender and about the original
 image processed. The body viewpoint and originating dataset can be linearly decoded from the embedding.
 \item We show that identity comparisons made 
 in the principal component space of deep network embeddings are consistently more accurate than those made on the raw embeddings.
  \item We show that it is possible to improve body identification accuracy by deleting parts of a network's latent space. We propose a technique to remove subspaces that are not relevant for identification in real applications.

\end{itemize}

\subsection{Background and Previous Work}
\subsubsection{Faces, bodies, and whole people in re-identification} 
Face and body recognition are commonly treated as separate problems in computational vision. Reasons for this
separation include differences in the expectation of uniqueness
of faces versus bodies for identification, as well as differences in the quality and resolution of the image needed to 
recognize a face versus a body.  In the real world, however, faces and bodies are seen together. As with humans,
body identification models learn individual
identities from images of whole people in the natural world. 
In the context of computation, 
body detection algorithms begin with an image of the whole person,
including the face and head region. 
Thus,
there is clear potential for the face/head to contribute
to the performance of these algorithms.

Motivated by privacy concerns in re-identification,
work to pinpoint the importance of face information in re-identification found only a minimal performance decrease when faces were blurred in the input image \cite{dietlmeier2021important}. However, five of the six datasets tested in that work had no clothing
change. It is possible, therefore, that clothing 
constancy compensated for the absence of a face.
Here we re-examine the role of the face in
long-term body identification algorithms tested with clothing-change
datasets. 


\subsubsection{Person and image attributes retained in body identification} 
Face representations (embeddings) generated by deep networks trained to identify faces retain detailed information that is not useful
for identification.
Specifically, they retain information about 
``instances of encounter''
in the form of image detail (e.g., viewpoint, illumination) \cite{parde2017face,hill2019deep,KRIZAJ2024107941,terhorst2020beyond}, expression \cite{colon_castillo_otoole_2021}, and 
appearance attributes (e.g., glasses, facial hair) \cite{terhorst2020beyond}).
Demographic categories, including gender, age, and race, are also retained \cite{dhar2020,hill2019deep}. These attributes can be ``read out'' from the DNN-generated face code with simple linear networks \cite{parde2017face,dhar2020,hill2019deep,parde2021closing}.
This type  of ``semantic'' information in the latent space
fuels the image-editing
capabilities \cite{shen2020interpreting} inherent in face-trained generative adversarial 
networks (GANS) \cite{goodfellow2020generative}.  

Semantic and image-based codes in the latent space of
body identification networks have not been investigated. As for faces, we might expect coding of demographic information such as gender,
as well as information based on image properties
that result from camera angle and environment.
We explore whether these semantic and image-based
attributes can be classified from 
body image embeddings.

\subsubsection{Subspace representations can improve identification} 
Applications of person identification networks often rely on raw embeddings. However, prior to the deep network era, modifying image representations to improve identification performance was not uncommon. Numerous applications in early facial recognition showed that subspaces of facial image representations yield features that are more effective at identification than complete representations (e.g., \cite{OToole1993}, \cite{Etemad:97}, \cite{Zhao1998}). Dimensionality reduction  in engineering
 nearly always refers to the deletion of PCs that explain small amounts of variance. However, 
 information useful for uniquely identifying a person should be features that  {\it are not shared
 with other identities} (i.e. information that explains
 small amounts of variance in a set of faces).  In older 
 image-based face
recognition models, deletion of PCs 
that explain larger amounts of variance 
increased identifcation accuracy \cite{OToole1993}. We explore this principle with deep network body identification
models. At a more basic level, 
work prior to deep networks showed that 
performing a simple PCA on image representations increased retrieval performance \cite{10.1007/978-3-642-33709-3_55}. Other  modern work in person identification 
also uses representation subspaces for identification. 
For example, by constructing multi-modal representations,
it is possible to find subspaces that improve identity discrimination \cite{10536789}. Here, we consider the question of whether simple, non-augmented representation transformations can  boost identification performance for deep networks trained for person re-identification. 



\section{METHODS}
We focus 
on four models. 
Two of these are ResNet-based CNNs \cite{he2015deepresiduallearningimage} and
two are Vision Transformers \cite{dosovitskiy2021imageworth16x16words,liu2021swintransformerhierarchicalvision}. 
In what follows, we describe the
model backbones and training.  The core training 
data (Table \ref{training_datasets}) and training methods described for the BIDDS model were used for all models,
except where noted.  All models 
were fit with a custom classification head mapped to a 2048-dimensional output space.

\subsection{Models and Training}
\subsubsection{Body Identification with Diverse Datasets (BIDDS)} \label{BIDDS_description}
The BIDDS model was built on a Vision Transformer architecture, using a ViT-B/16 variant pre-trained on ImageNet-1k. The core model processes $224\times224$ pixel images with patch size 16. 
All models were 
trained to map images of bodies to identities (see Table \ref{training_datasets}). 
Traditional re-id datasets like Market1501 \cite{zheng2015scalable} and MSMT17 \cite{wei2018person} are
included, as well as  challenging long-term re-identification clothes-change datasets like DeepChange \cite{xu2023deepchange} and the BRIAR Research Dataset (BRS) \cite{cornett2023expanding}. The training data include a range of distances, camera angles, and weather conditions. 

Hard triplet loss with negative mining \cite{hermans2017defense} was used for training. This operates on image triplets: an anchor image, a positive sample (same identity), and a negative sample (different identity). The loss calculation measures the Euclidean distances between the anchor and positive samples and between the anchor and negative samples. We selected the most challenging negative samples (i.e., those closest to the anchor in the embedding space) within each batch. This hard negative mining encourages the model to learn features that effectively differentiate between similar body shapes.
We used 
the Adam optimizer and incorporated dynamic sampling, whereby triplet selection is adapted based on the current state of the embeddings. This ensures that the model continuously encounters challenging examples throughout training \cite{kingma2017adammethodstochasticoptimization}. The training process employs a low learning rate ($10^{-5}$) and weight decay ($10^{-6}$) to prevent over-fitting while maintaining  stability.

Following core training, the BIDDS model was fine-tuned on the BRS1--5 datasets (cf. \cite{cornett2023expanding}), increasing input resolution to $384\times384$ pixels. 

\subsubsection{Swin Transformer Body Identification with Diverse Datasets (Swin-BIDDS)}
Swin-BIDDS utilizes the hierarchical vision transformer, which 
uses shifted windows \cite{liu2021swin}. 
This  transformer is more efficient, because it
  limits self-attention computation to non-overlapping local windows,  while supporting cross-window connections. The hierarchical structure 
progressively merges patches and is 
 well-suited to
 modeling at various scales.
 We used an ImageNet-1k pretrained version
 of the model.  The model was core trained (see Section \ref{BIDDS_description}) with 
 input resolution at $384\times384$ pixels.
Swin-BIDDS fine-tuning was identical to BIDDS fine-tuning
(see Section \ref{BIDDS_description}).

\subsubsection{Linguistic Core ResNet Identity Model (LCRIM)}
LCRIM incorporates human body descriptors into its 
training.\footnote{Earlier versions of LCRIM and NLCRIM were trained with substantially less data and without
triplet loss hard negative mining \cite{myers2023recognizing}.} A ResNet-50 model pretrained on ImageNet-1k \cite{russakovsky2015imagenet} was used as a base. The base was augmented with an encoder/decoder structure that maps to a linguistic feature space
before the final identification layers. The encoder pathway compresses the representation (2048 → 512 → 64 → 16), while the decoder pathway (16 → 24 → 30) reconstructs linguistic body attributes.  
LCRIM was pretrained to produce body descriptors on the HumanID \cite{OToole_2005} and MEVA
\cite{Corona_2021_WACV} datasets. HumanID consists of videos of 297 identities filmed from multiple angles, viewpoints, and lighting conditions. Subjects are filmed wearing different clothing sets. The MEVA dataset, comprising over 9,300 hours of video across varied activities and scenarios, contributed an additional 158 identities. 
Identities were annotated by 20 subjects for 30 human descriptors (e.g., broad shoulders, feminine, hourglass, long legs, pear-shaped, petite; for a full list, see \cite{myers2023recognizing}). 
Images from these datasets were used to train LCRIM's initial ability to map between visual features and linguistic body descriptions.

The linguistic-core training was followed by a core training identical to that described in Section \ref{BIDDS_description}. There was no additional fine-tuning.

\subsubsection{Non-Linguistic Core ResNet Identity Model (NLCRIM)} NLCRIM was identical to LCRIM, but with no
linguistic training.



\begin{table}
\center
\caption{Training Datasets}
\label{training_datasets}
\begin{tabular}{|p{2.6cm}||p{1.25cm}||p{1.25cm}||p{1.5cm}|}
\hline
\textbf{Dataset} & \textbf{Images} & \textbf{IDs} & \textbf{Clothes Change} \\
\hline
UAV-Human \cite{9578530} & 41,290 & 119  & no \\\hline
MSMT17 \cite{wei2018person} & 29,204 & 930  & no \\\hline
Market1501 \cite{zheng2015scalable} & 17,874  & 1,170  & no \\\hline
MARS \cite{zheng2016mars} & 509,914 & 625  & no  \\\hline
STR-BRC \cite{cornett2023expanding} & 156,688 & 224  & yes  \\\hline
P-DESTRE \cite{kumar2020p} & 214,950  &  124  & no \\\hline
PRCC \cite{yang2019person} & 17,896 & 150  & yes \\\hline
DeepChange \cite{xu2023deepchange} & 28,1731 &  451  & yes\\\hline
BRS 1--5 \cite{cornett2023expanding} & 697,348  & 995  & yes  \\\hline
\hline
\textbf{Total Training} &  1,966,895 & 4,788 & \\\hline
\end{tabular}
\vskip -0.5cm
\end{table}

\subsection{Test Datasets}
Three clothes-change data sets were used for the experiments.
1.) {\it Person Reidentification by Countour Sketch under Moderate Clothing Change (PRCC)}  \cite{yang2019person}.
Images were captured across three cameras. Two cameras captured images in the same clothing set in different settings. The third camera captured images in different clothing sets on different days. 
2.) {\it DeepChange} \cite{xu2023deepchange}. This set consists of 178k bounding boxes of 1.1k identities. Images are captured across 17 cameras from real-world surveillance systems and include changes in clothing as well as hairstyle. 
3.) {\it HumanID} \cite{OToole_2005}. The HumanID dataset consists of videos of 297 identities filmed from multiple angles, viewpoints, and illumination conditions. We analyzed 94 of these identities. This subset had metadata for analysis of the camera angle.  

\begin{figure}[thpb]
      \centering
\includegraphics[scale = .5]{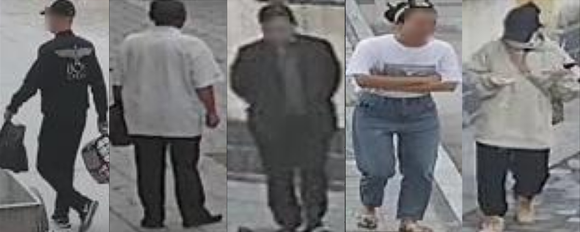}
      \vskip - 0.35cm
      \caption{Sample images from the DeepChange dataset. Subject faces are blurred when visible according to DeepChange publication guidelines.}
      \label{deep_change_samples}
      \vskip -0.5cm
   \end{figure}
   
\section{Experiments and Results}
\subsection{Experiment 1---Face Contribution to Body Identification} In Part 1 of the experiment, we tested the models using body images
    with the face  obscured. In Part 2, we tested the models using only the face. 
 Bounding boxes for faces were found for the PRCC and HumanID dataset using a Multi-task Cascaded Convolutional Neural Network (MTCNN) trained to detect faces \cite{Zhang_2016}. 
 Images in
 which the detector failed to find a face were included in Part 1 and discarded for Part 2. The DeepChange dataset was excluded from both parts of Experiment 1, because faces were undetected in $\approx97\%$ of the images, and because no faces were detected for certain identities, making face-based analyses impractical. 
Unless specified, input images were normalized using ImageNet means and standard deviations. Images were resized to 224$\times$224 for LCRIM and NLCRIM and 384$\times$384 for BIDDS and Swin-BIDDS.
\subsubsection{Part 1---Face Obscured} Images in which faces were found were edited to place a black box over the face. The remaining images were left unchanged, under the assumption that no (or limited) facial information was available in these images. 
We calculated performance metrics 
(area under the ROC curve, AUC; mean average precision, mAP; Rank 1 and Rank 20 identification) on the modified datasets (black boxes drawn over faces found by MTCNN) and unmodified datasets (same images but with no transformations applied).
\begin{figure}[thpb]
      \centering
       \includegraphics[scale=.7]{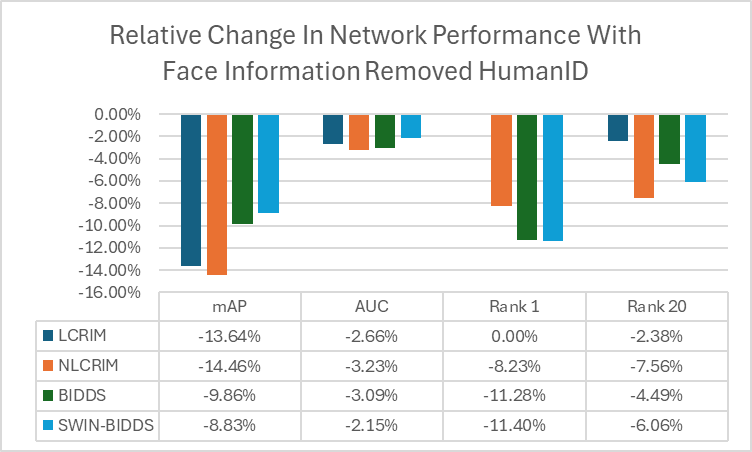}
      \includegraphics[scale = .7]{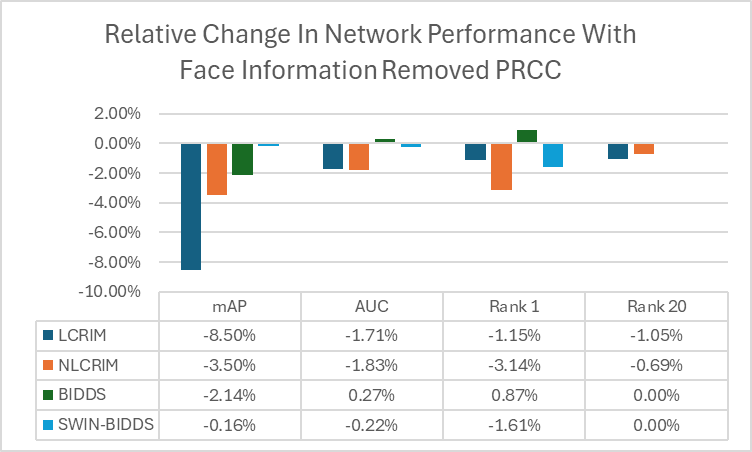}
      \vskip - 0.25cm
      \caption{The performance of the four body identification networks decreases when facial information is obscured. }
      \label{table:face_remove}
      \vskip - 0.70cm
   \end{figure}

 Figure  \ref{table:face_remove} shows that obscuring the face degrades identification accuracy  for all models in both datasets,
 on most metrics. There was one exception: accuracy for the BIDDS model decreased
only for mAP on the PRCC dataset.
Performance dropped more on the challenging HumanID dataset than on the less challenging PRCC dataset. This suggests that the face may contribute relatively more to identification when the image quality overall is poor.
The baseline accuracy of the four models was also a factor in performance when the face was obscured. The more accurate models overall (BIDDs, Swin-BIDDS) appeared to be more robust to the removal of facial information 
 than the less accurate models (LCRIM, NLCRIM) (see Fig. \ref{table:face_remove}).  Without the face, metrics of matching accuracy such as Rank-1 and mAP declined  more than AUC and Rank-20. This suggests that the face plays a key role in performing identity matching. These results demonstrate that identity-trained
 long-term body re-identification models utilize facial information when it is available---especially on challenging identification tasks.

\subsubsection{Part 2---Face Only}
Next, we tested the body-trained networks
on cropped faces and 
compared identification
accuracy with that of a high-performing
facial identification network. To measure the baseline identification accuracy for the cropped faces (Face Baseline), we used a Controllable and Guided Face Synthesis for Unconstrained Face Recognition network, built on top of an ArcFace recognition module 
\cite{liu2022controllableguidedfacesynthesis,Deng_2022}. 
This  network was selected due to its strong performance on difficult unconstrained face identification datasets, such as IJB-B and IJB-C \cite{maze2018iarpa}.  Consistent with the specified training procedures  (\cite{liu2022controllableguidedfacesynthesis,Deng_2022}), 
images were normalized by mean = $[.5, .5, .5]$ and standard deviation = $[.5, .5, .5]$ and resized to 112x112 before being fed to the Face Baseline.
 
\begin{figure}[thpb]
     \includegraphics[scale = .57]{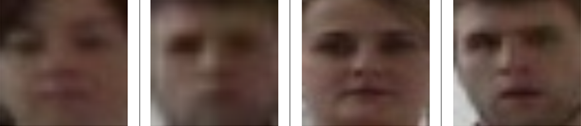}
    \vskip - 0.25cm
    \caption{Sample cropped face images from the body dataset \cite{OToole_2005}. All subjects consented
    to publication.}
          \vskip - 0.45cm
      \label{face_only_image_examples}
   \end{figure}

 From the HumanID and PRCC datasets, two very challenging face datasets were constructed by cropping faces detected by MTCNN and replacing the original images with the cropped face images. Images in which a face was not found were discarded. 
 The body identification networks were tested on these datasets, consisting only of cropped faces. Figure \ref{face_only_image_examples} shows example images.
 
\begin{figure}[thpb]
      \centering
     \includegraphics[scale = .65]{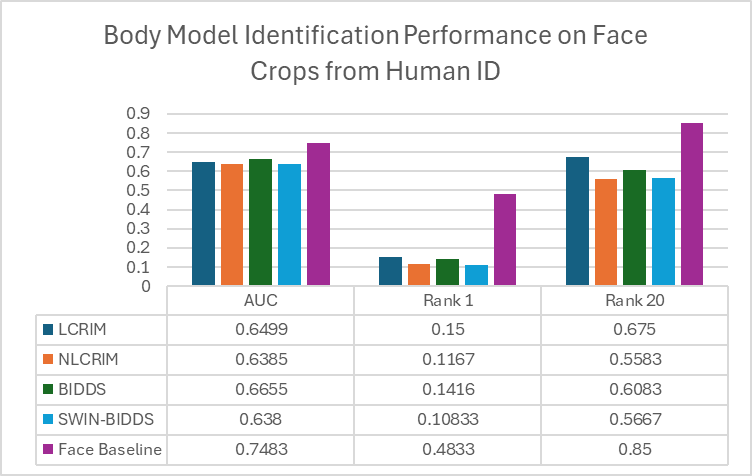}
      \includegraphics[scale = .65]{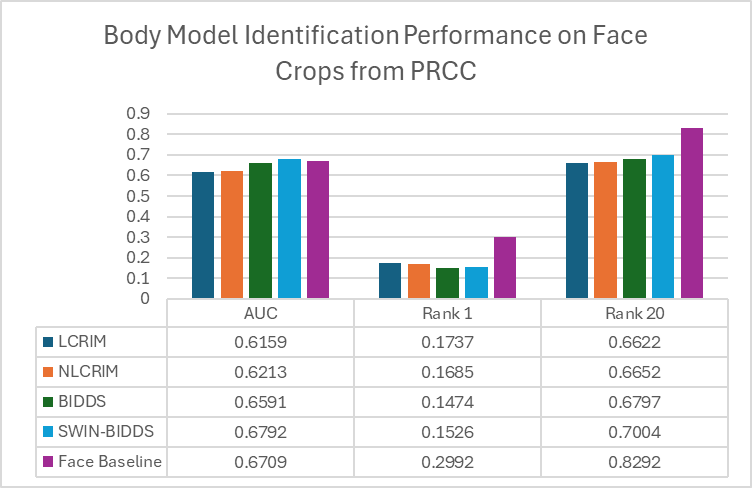}
      \vskip - 0.35cm
    \caption{Body networks learn residual information about face in their training. These networks can perform unconstrained face identification tasks to some degree.}
          \vskip - 0.25cm
      \label{face_only_plots}
   \end{figure}

The plots in Fig. \ref{face_only_plots} show that body re-identification networks learn useful information
about faces. As expected, identification of these low quality faces from the body identification networks, overall, 
was far from perfect. However, relative to the face 
baseline network, performance is unexpectedly strong. 
For example, on the PRCC face data set, the AUC of the BIDDS model actually exceeded the baseline of the face-trained network. In all cases, AUC and rank 20 performance 
show that the body networks recognize
faces with some degree of accuracy. 
As expected, rank 1 performance for all models, even
the face baseline, is low. This level of performance is appropriate
given the image quality.
Consistent with the  face-obscured results, these findings suggest that body-trained identification models utilize information in the face.

Given the low quality of facial information in the HumanID and PRCC body-datasets,
 we constructed a test set from the popular (better quality) Labeled Faces in the Wild (LFW)  face dataset \cite{huang2008labeled}. 
Utilizing all identities that contain more than one image per identity, the test set contained 9,164 images of 1,680 identities, evenly split over a randomly assigned probe and gallery set.   
\begin{figure}[thpb]
      \centering
\includegraphics[scale = .65]{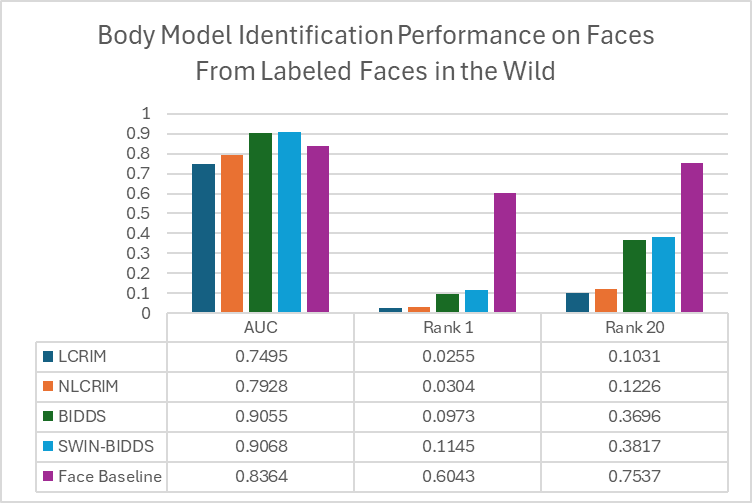}
      \vskip - 0.35cm
      \caption{Body Networks learn residual information about faces in their training. This information generalizes to high quality faces.}
      \label{figurelabel}
    \vskip - 0.35cm
   \end{figure}
On this higher-quality face test, the body-identification models still perform better than expected  with BIDDS and Swin-BIDDS surpassing the face baseline AUC. However, all body models perform substantially worse than the face baseline on retrieval tasks. 

In summary, when the face was obscured, the stronger networks were more robust to the removal of face information. These  networks also identified  faces more accurately. Combined,
the results suggest that the stronger person-identification networks use facial information more effectively 
when it is available, and depend less on the face
 when it is not available.

\subsection{Experiment 2---Gender and Image Attribute Retention}
In Part 1, we examine whether gender information is retained
in body embeddings from the body identification networks.
In Part 2, we examine the retention of  image-specific  information 
(viewpoint, dataset).
\subsubsection{Part 1---Gender}
 
Gender (male, female) was labeled manually for the PRCC, HumanID, and DeepChange datasets.  Half of the identities from each dataset were randomly pooled into a single training set. The remaining identities comprised the test set. A logistic regression model was trained to predict gender from the image embeddings in the training set. The model was tested using the held-out image embeddings. 

\begin{figure}[thpb]
      \centering
\includegraphics[scale = .7]{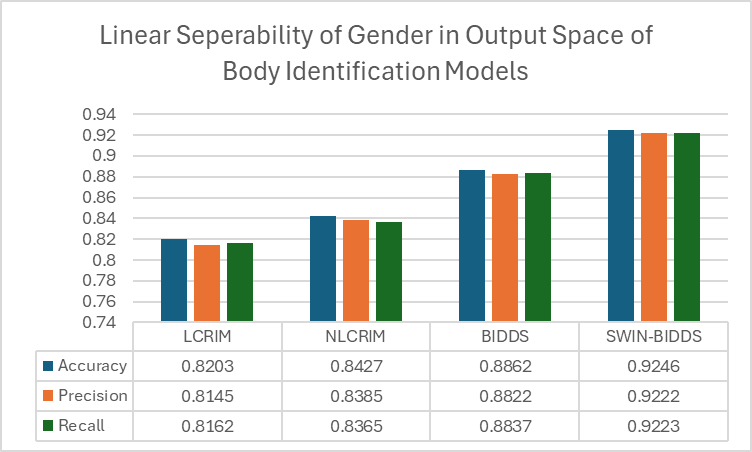}
      \vskip - 0.25cm
      \caption{Body network embeddings retain gender information which can be read out using linear models.}
      \label{gender_prediction}
      \vskip - 0.55cm
   \end{figure}

Figure \ref{gender_prediction} shows that 
the network embeddings retain information about gender. This is not entirely surprising, as gender serves as a useful discriminator when describing human bodies. Similar results have been reported for the analysis of face identification networks \cite{hill2019deep,parde2017face,parde2021closing}. This result is promising for  future work with body-based GANS, which could utilize this sematic information as a basis for image editing. 

\subsubsection{Part 2---Image-based information}

We begin with the viewpoint, which is well-controlled and annotated in the HumanID dataset. This dataset provides body images from two viewpoints. These correspond to a person walking directly towards the camera (0$\degree$ yaw) and walking perpendicular to the camera (90$\degree$ yaw). From each viewpoint, a sequence of images was captured. These images were available for each person with two different clothing sets. A training set composed of embeddings from half of the images in the HumanID set was constructed; the remaining images were held out for testing. Again, logistic regression models were trained to predict the person's yaw (0$\degree$ vs. 90$\degree$) in the image. 

\begin{figure}[thpb]
      \centering
      \includegraphics[scale = .7]{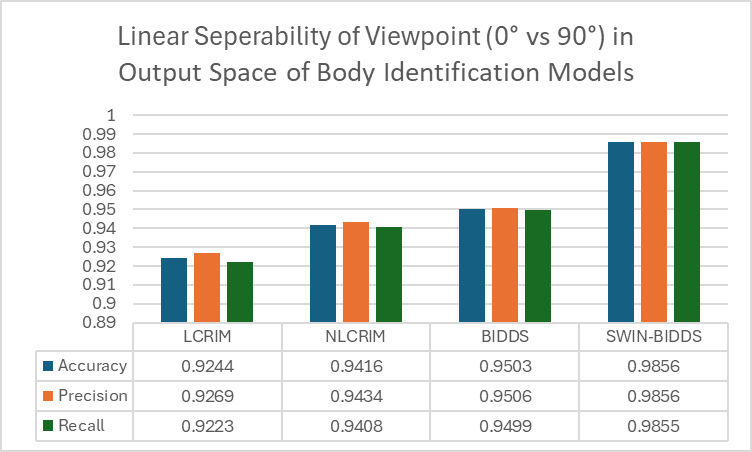}
      \vskip - 0.25cm
      \caption{Body network embeddings retain  image viewpoint information which can be read out using linear models.}
      \label{viewpoint_prediction}
            \vskip - .35cm
   \end{figure}

Figure \ref{viewpoint_prediction} shows the results on the test set.  
All of the models retain information about the yaw of the person in the image. 
This finding is consistent with  previous analyses for face networks \cite{hill2019deep}.
Surprisingly, predictions of camera angle and gender demonstrate a relationship with model accuracy. Paradoxically, stronger person identification models seem to retain more non-identity information in their embeddings than weaker models.

Next, we asked whether the dataset of origin of the image could be predicted from
the image embedding. Half of the identities from PRCC, HumanID, and DeepChange were pooled into a training set. The remaining half of identities served as the test set. A logistic regression model was trained to predict dataset of origin. The embeddings corresponding to each image for each identity were input
and the dataset of origin label was predicted.  

\begin{figure}[thpb]
      \centering
\includegraphics[scale = .7]{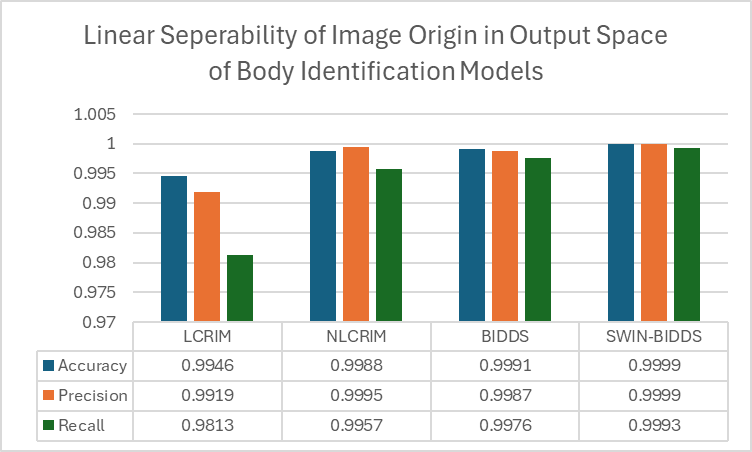}
      \vskip - 0.25cm
      \caption{Body network embeddings retain information pertaining to the origin of an image which can be read out using linear models.}
      \label{dataset_of_origin_prediction}
      \vskip - 0.35cm
   \end{figure}

Figure \ref{dataset_of_origin_prediction} shows that
all models retain substantial information about the dataset from which an
image was taken. This indicates that presumably trivial aspects of image
quality, imaging conditions, or camera are retained in the embedding. Further experimentation is required to uncover exactly what information is retained to enable this prediction. 
Regardless, there are obvious potential applications in forensics and
security for being able to trace image
characteristics from a deep network embedding.

\subsection{Experiment 3---Principal Component Analysis (PCA) and Dimensionality Reduction}
\subsubsection{Part 1---Identification in PC Space}\label{PCA_ID}
In standard body identification tasks,  image embeddings
for gallery identities are compared to  embeddings
for  probe identities. 
Here, we applied PCA  to all gallery image embeddings. The original gallery embeddings $X_g$ were then projected into the  PC space yielding $X_g^p$. The probe set $X_p$ was likewise projected into the PC space yielding $X_p^p$. Finally, performance metrics were calculated on $X_g^p$ and $X_p^p$. 


\begin{figure}[tb]
    \centering
    \includegraphics[scale=0.44]{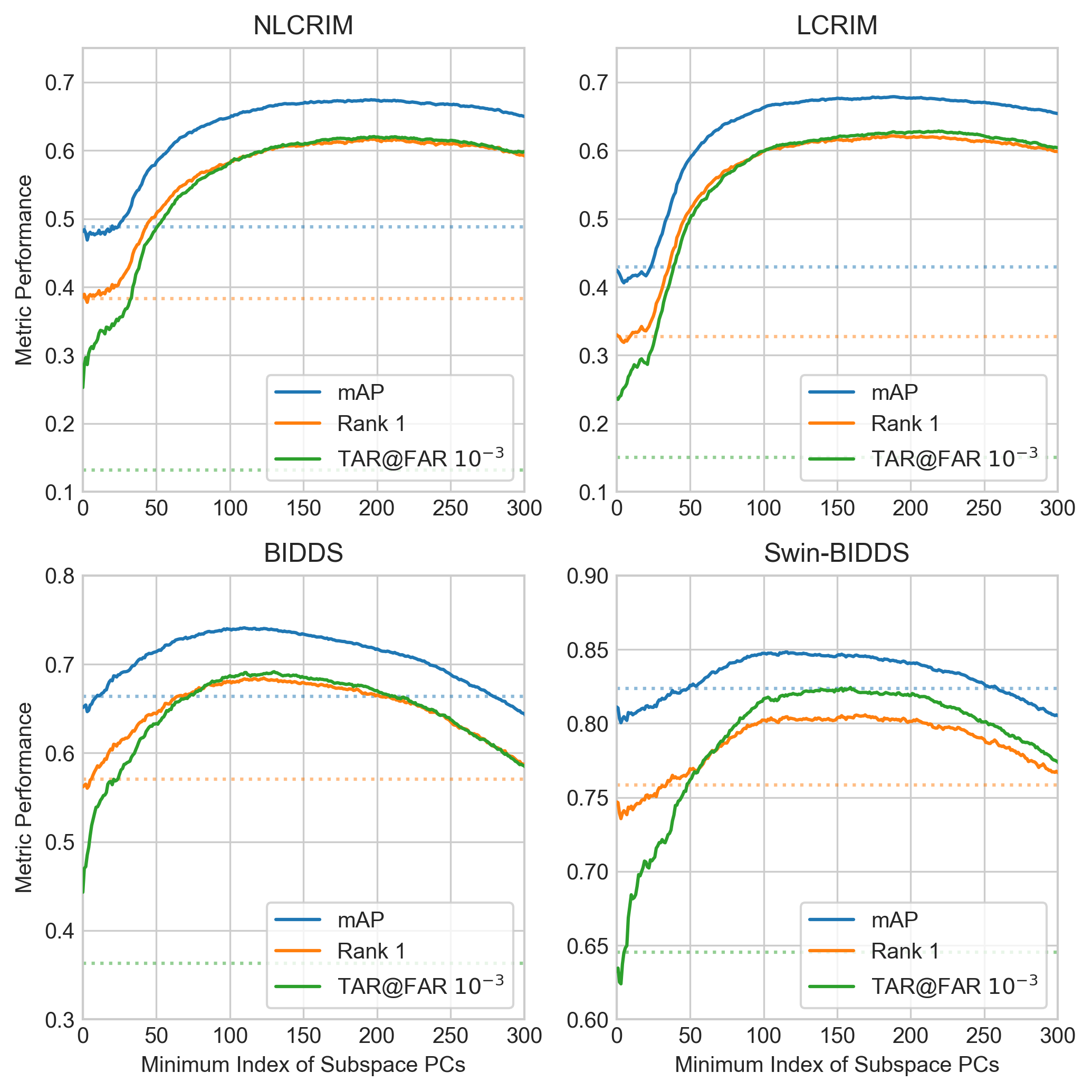}
    \vskip - 0.35cm
      \caption{{\it DeepChange Dataset.} Using an Oracle Algorithm to excise early principal components can lead to substantial boosts in Rank1, and TAR@FAR$10^{-3}$, and mAP with some degradation in AUC (not shown). Horizontal lines indicate metric baselines from raw embeddings.\protect\footnotemark[2]}
      \label{PCA_optimal_subspace_-_DC}
      \vskip - 0.45cm
   \end{figure}

\begin{figure}[tb]
    \centering
    \includegraphics[scale=0.44]{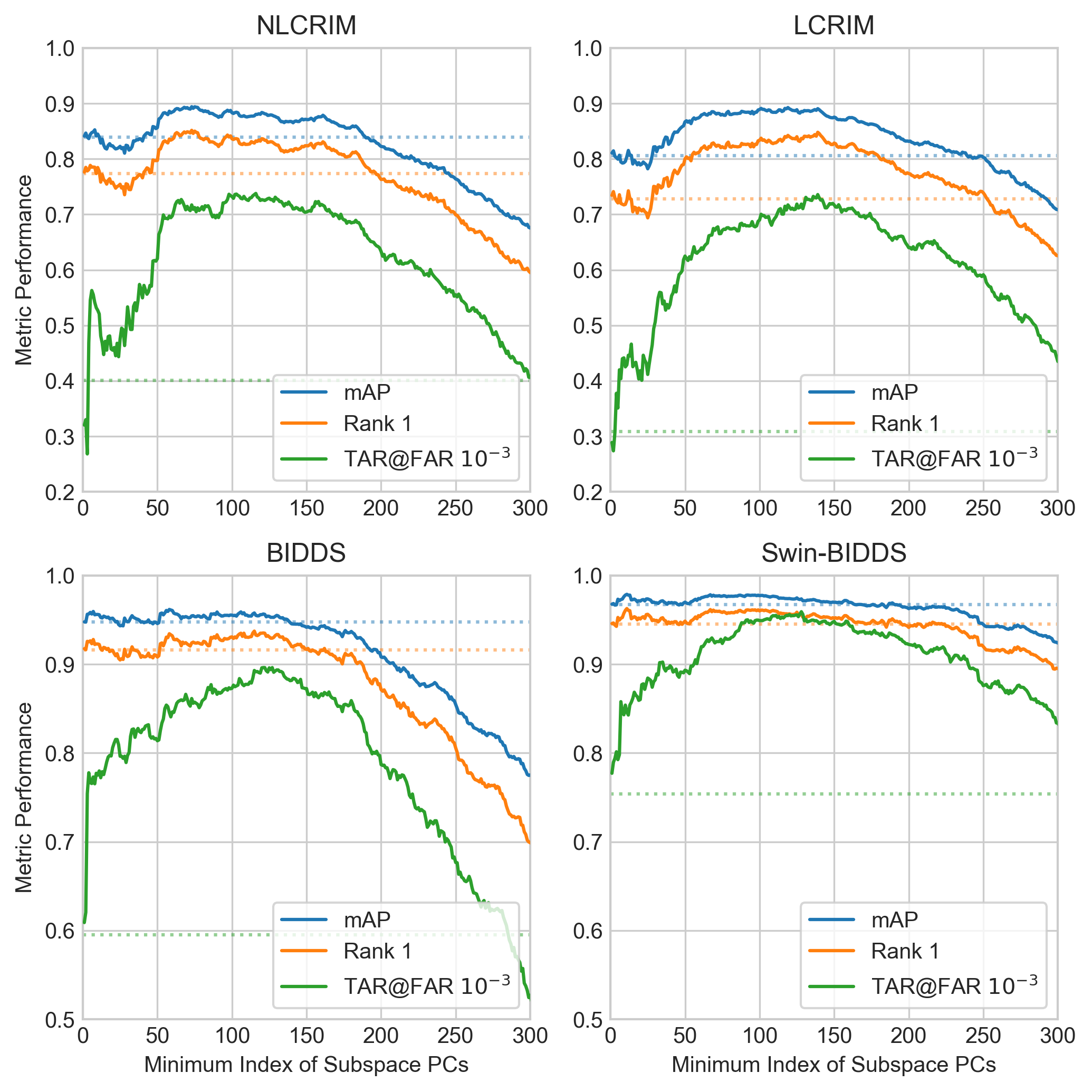}
    \vskip - 0.35cm
      \caption{{\it PRCC Dataset}. Using an Oracle Algorithm to excise early principal components can lead to substantial boosts in Rank1, TAR@FAR$10^{-3}$, and mAP with some degradation in AUC (not shown). Horizontal lines indicate metric baselines from raw embeddings.\protect\footnotemark[2]}
      \label{PCA optimal subspace - PRCC}
      \vskip - 0.45cm
   \end{figure}

\begin{figure}[tb]
    \centering
    \includegraphics[scale=0.44]{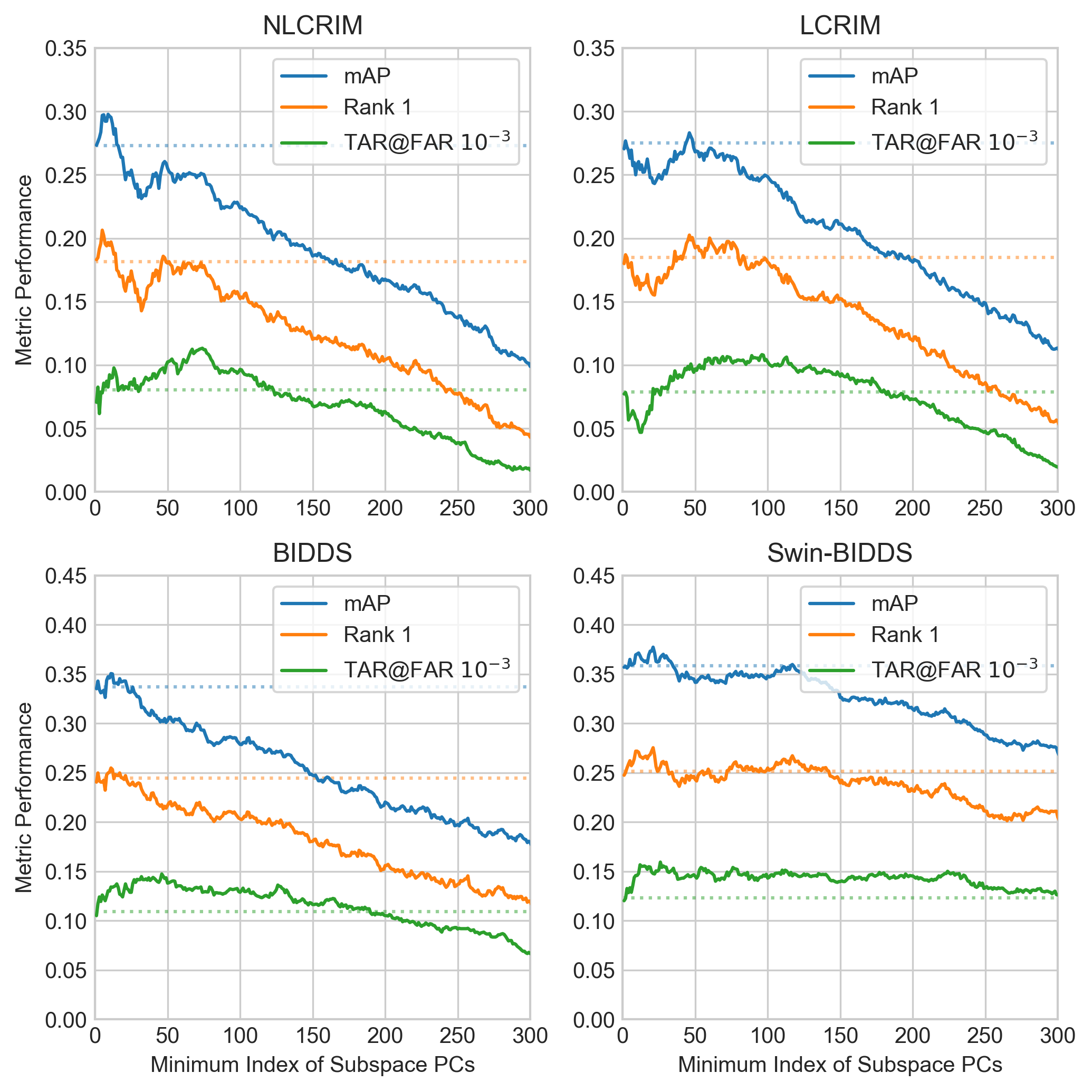}
    \vskip - 0.35cm
    \caption{{\it HumanID Dataset}. Using an Oracle Algorithm to excise early principal components can lead to modest boosts in Rank1, TAR@FAR$10^{-3}$, and mAP with some degradation in AUC (not shown). Horizontal lines indicate metric baselines from raw embeddings.\protect\footnotemark[2]}
    \label{PCA optimal subspace - HID}
    \vskip -0.45cm
\end{figure}

\footnotetext[2]{AUC and mAP metrics shown here are not comparable to those in Table \ref{tab:PCA_applied}, because AUC is calculated on a templated gallery set. This distinction is made because this figure describes an approach that would be used in a re-identification system which would rely on templates.}

\begin{table}[ht]
\centering
\begin{tabular}{lcccccccccc}
\textbf{Model} & \textbf{PCA} & \textbf{Dataset} & \textbf{MaP} & \textbf{AUC} & \textbf{$\Delta$AUC} & \textbf{$\Delta$MaP} \\
\hline
LCRIM  & No  & DC & .1125 & .7223 & - & - \\
LCRIM  & Yes & DC & .1189 & .7385 & \textcolor{green}{↑}2.24$\%$ & \textcolor{green}{↑}5.69$\%$ \\
NLCRIM & No  & DC & .1537 & .7462 & - & - \\
NLCRIM & Yes & DC & .1625 & .7695 & \textcolor{green}{↑}3.12$\%$ & \textcolor{green}{↑}5.73$\%$ \\
BIDDS    & No  & DC & .2520 & .7905 & - & - \\
BIDDS    & Yes & DC & .2583 & .8062 & \textcolor{green}{↑}1.99$\%$ & \textcolor{green}{↑}2.50$\%$ \\
SWIN-BIDDS  & No  & DC & .4325 & .9210 & - & - \\
SWIN-BIDDS   & Yes & DC & .4339 & .9153 & \textcolor{green}{↑}0.15$\%$ & \textcolor{red}{↓}1.32$\%$ \\
\hline
LCRIM  & No  & PRCC & .4213 & .8873 & - & - \\
LCRIM  & Yes & PRCC & .4646 & .8939 & \textcolor{green}{↑}0.74$\%$ & \textcolor{green}{↑}10.28$\%$ \\
NLCRIM & No  & PRCC & .4882 & .9030 & - & - \\
NLCRIM & Yes & PRCC & .5093 & .9128 & \textcolor{green}{↑}1.09$\%$ & \textcolor{green}{↑}4.32$\%$ \\
BIDDS    & No  & PRCC & .6578 & .9196 & - & - \\
BIDDS    & Yes & PRCC & .6817 & .9417 & \textcolor{green}{↑}2.40$\%$ & \textcolor{green}{↑}3.63$\%$ \\
SWIN-BIDDS   & No  & PRCC & .7020 & .9390 & - & - \\
SWIN-BIDDS   & Yes & PRCC & .7147 & .9495 & \textcolor{green}{↑}1.35$\%$ & \textcolor{green}{↑}1.50$\%$ \\
\hline
LCRIM  & No  & H ID & .1261 & .6963 & - & - \\
LCRIM  & Yes & H ID & .1557 & .7326 & \textcolor{green}{↑}5.21$\%$ & \textcolor{green}{↑}23.47$\%$ \\
NLCRIM & No  & H ID & .1342 & .7332 & - & - \\
NLCRIM & Yes & H ID & .1574 & .7615 & \textcolor{green}{↑}3.72$\%$ & \textcolor{green}{↑}17.28$\%$ \\
BIDDS    & No  & H ID & .1907 & .7706 & - & - \\
BIDDS    & Yes & H ID & .2192 & .7910 & \textcolor{green}{↑}1.20$\%$ & \textcolor{green}{↑}14.94$\%$ \\
SWIN-BIDDS   & No  & H ID & .2253 & .8009 & - & - \\
SWIN-BIDDS   & Yes & H ID & .2566 & .8192 & \textcolor{green}{↑}2.28$\%$ & \textcolor{green}{↑}13.90$\%$ \\
\end{tabular}
\caption{Performance comparison across DeepChange (DC), PRCC, and HumanID (H ID) datasets with and without PCA applied.}
\label{tab:PCA_applied}
\vskip - 1.1cm
\end{table}

Table  \ref{tab:PCA_applied} shows that
identification accuracy improves in the PCA space for all metrics and models, across the three datasets. There was only one exception---AUC for the Swin-BIDDS model on the DeepChange dataset declined slightly. 

Performance boosts on average were modest, but 
highly consistent, and sometimes substantial.  
This shows that a computationally inexpensive and well-known technique that requires no additional training 
can be used to improve body identification accuracy. 

\subsubsection{ Part 2---Selective Dimensionality Reduction in PC Space} 
First, we show that reducing the dimensionality of the PC space {\it can} lead to substantial performance increases for our person-identification models.  Dimensionality reduction
 was implemented by deleting the PCs that 
{\it explain the most variance} (cf. \cite{OToole1993})
(i.e., delete $PC_1$, then $PC_{1-2}$ ... $PC_{1-n}$). 
We begin with an ``oracle simulation''  that uses both probe and gallery items to implement dimensionality reduction.
We then propose and test a method for dimensionality
reduction that operates only on the gallery items, and
can thereby be used in real-world applications. We show that this method achieves an appreciable performance increase.   

\begin{figure}[thpb]
      \centering
       \includegraphics[scale = .67]{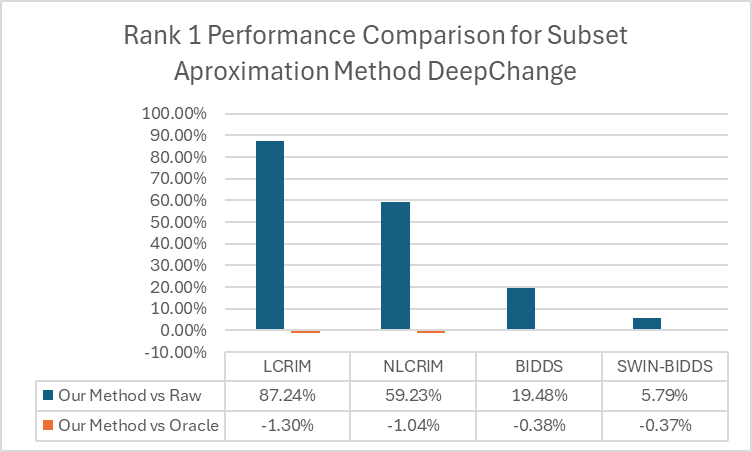}
\includegraphics[scale = .67]{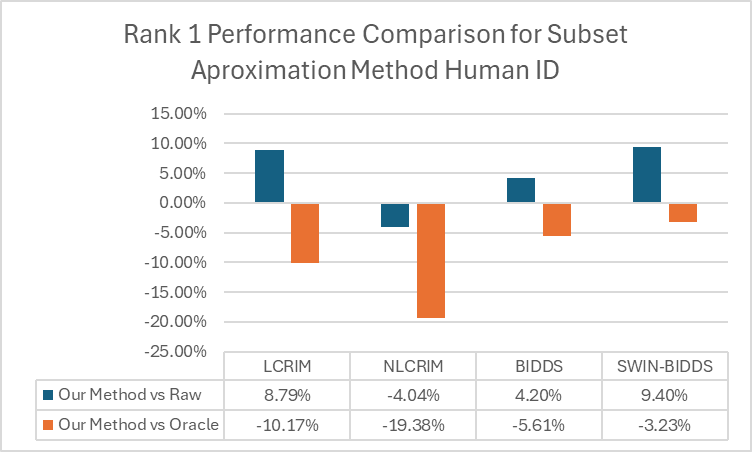}
\includegraphics[scale = .60]{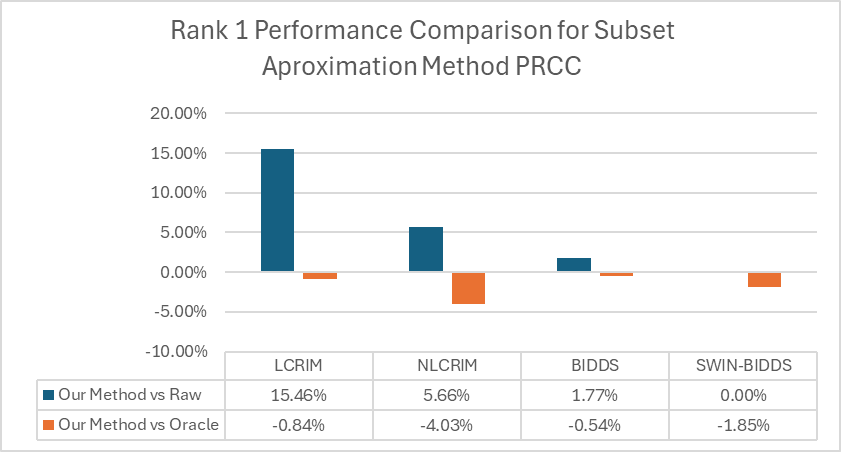}
\vskip - 0.35cm
      \caption{Better subsets for rank 1 retrieval can be approximated with only the gallery set. Our method refers to performance achieved by algorithm \ref{alg:general}, oracle refers to the optimal performance that can be achieved using algorithm \ref{alg:Optimal}, and raw refers to performance achieved by using unmodified network embeddings.}
      \label{PCA technique}
      \vskip -0.45cm
   \end{figure}
The oracle method is described in Algorithm \ref{alg:Optimal}. We used network embeddings for each image in each dataset's gallery ($X_{g,DeepChange}$, $X_{g,PRCC}$, and $X_{g,HumanID}$). Because $X_{g,PRCC}$, $X_{g,HumanID}$ contain a small number of identities relative to the embedding space dimensionalities (71 and 94 respectively vs. 2048), a fourth gallery set was constructed $X_{g,all}$ by combining the gallery embeddings of the three datasets. This larger gallery supports the construction of informative PC representations. 

Figs. \ref{PCA_optimal_subspace_-_DC} and \ref{PCA optimal subspace - PRCC} show the oracle results for DeepChange and PRCC,
respectively. At a general level, the metrics follow an inverted U-shaped function  as earlier PCs are progressively
excised. Specifically, Rank 1, mAP, and 
TAR@FAR$10^{-3}$ increase substantially
when earlier PCs are excised before classification. This peaks and
declines when very large numbers of PCs are eliminated. 
For the HumanID dataset, Fig. \ref{PCA optimal subspace - HID} shows that Rank 1, mAP, and 
TAR@FAR$10^{-3}$ increase modestly before declining. 
Moreover, the performance peak occurs in much earlier principal components than for DeepChange and PRCC. This difference may arise from the difficult and small gallery 
available in HumanID. 
In summary, we show that across DeepChange, PRCC, and HumanID, eliminating early PCs {\it can} modestly boost and sometimes substantially boost multiple performance metrics.

Next, we propose and test a method for finding effective PC subspaces  that can  improve retrieval accuracy, based only on operations that use  a pre-defined gallery set (see Algorithm \ref{alg:general}). We test this on the three original (smaller) datasets. Here we focus on Rank 1 retrieval because it is most reflective of performance changes on small datasets like PRCC and HumanID. We compare our method to the use of the raw embeddings
and to the oracle-selected embeddings.

Figure \ref{PCA technique} shows that Rank 1 retrieval increases for the models on all datasets, with one exception: NLCRIM on the HumanID Dataset when the PC subspace is utilized in place of unmodified embeddings. 
Performance increases vary according to dataset and model. On DeepChange, some models realize a near than 90$\%$ increase in Rank 1 performance. Comparatevly, on HumanID the greatest increase in Rank 1 performance is $10\%$. 
Of note, the generally less performant models (e.g., LCRIM) 
show a larger benefit than the stronger models (BIDDS and SWIN-BIDDS). Also of note, Figure \ref{PCA technique} shows that 
when using only a pre-defined gallery set, 
the performance of Algorithm \ref{alg:general} often approaches that of the oracle. Thus, using only a predefined gallery set, Algorithm \ref{alg:general} can substantially boost rank 1 retrieval, often nearly as well as an oracle algorithm.  

In summary, excising early principal components from latent network representations consistently and often substantially boosts body identification network performance in both an oracle simulation and in a real world context.

\begin{algorithm}
\caption{Oracle Algorithm for Demonstrating the Effect of PC Subspaces on Identification Accuracy}\label{alg:Optimal}
\begin{algorithmic}[1]
\State \textbf{Input:} Network embeddings for Probe and Gallery sets $X_p$ and $X_g$
\State \textbf{Output:} PC subspaces and their corresponding identification metrics (mAP, Rank-1, TAR@FAR$10^{-3}$)


\State \textbf{Step 1: Construct Template Gallery ($X_{g,t}$):}
\For{each identity in $X_g$}
    \State Compute the average of all embeddings for the given identity
    \State Add the averaged embedding as a template into $X_{g,t}$
\EndFor


\State \textbf{Step 2: Compute PCA Basis:}
\State Perform PCA on $X_{g,t}$
\State Project $X_{g,t}$ and $X_p$ into the PCA basis, yielding $X_{g,t}^{PCA}$ and $X_p^{PCA}$


\State \textbf{Step 3: Order Principal Components:}
\State Rank the eigenvectors of $X_{g,t}^{PCA}$ in descending order of explained variance


\State \textbf{Step 4: Iterative Subspace Analysis:}
\For{$k = 1$ to total number of eigenvectors}
    \State Remove the $k$ largest eigenvectors (based on explained variance)
    \State Evaluate identification metrics (mAP, Rank-1, TAR@FAR$10^{-3}$) using the subspace of $X_{g,t}^{PCA}$ and $X_p^{PCA}$
\EndFor


\State \textbf{Return:} Return all evaluated subspaces and their corresponding identification metrics
\end{algorithmic}
\end{algorithm}

\begin{algorithm}
\caption{Subspace Approximation Method for Improving Accuracy}\label{alg:general}
\begin{algorithmic}[1]
\State \textbf{Input:} Network embeddings for Probe and Gallery sets $X_p$ and $X_g$
\State \textbf{Output:} Modified embeddings for retrieval tasks


\State \textbf{Step 1: Construct Template Gallery ($X_{g,t}$):}
\For{each identity in $X_g$}
    \State Compute the average of all embeddings for the given identity
    \State Add the averaged embedding as a template into $X_{g,t}$
\EndFor


\State \textbf{Step 2: Compute PCA Basis:}
\State Perform PCA on $X_{g,t}$
\State Project $X_{g,t}$, $X_p$, and $X_g$ into the PCA basis, resulting in $X_{g,t}^{PCA}$, $X_p^{PCA}$, and $X_g^{PCA}$


\State \textbf{Step 3: Subspace Search for Accuracy Improvement:}
\For{each principal component}
    \State Remove the principal component with the highest explained variance
    \State Evaluate Rank-1 performance between $X_{g,t}^{PCA}$ and $X_g^{PCA}$
\EndFor


\State \textbf{Step 4: Final Embedding Construction:}
\State Identify the principal components corresponding to the highest Rank-1 performance between $X_{g,t}^{PCA}$ and $X_g^{PCA}$
\State Remove these components from $X_{g,t}^{PCA}$ and $X_p^{PCA}$ to create $X_{g,t}^{Final}$ and $X_p^{Final}$


\State \textbf{Return:} $X_{g,t}^{Final}$ and $X_p^{Final}$

\end{algorithmic}
\end{algorithm}
\setlength{\textfloatsep}{0pt}

\addtolength{\textheight}{-3cm}   
\section{CONCLUSIONS AND FUTURE WORKS}

\subsection{Conclusions}
This work provides an extensive analysis of four 
long-term person re-identification models. We show that 
facial information contributes to the performance
of person re-identification models trained with whole-body images. Concomitantly, these models display a residual ability
to identify people using only the face. We demonstrate a number of parallels between body  and face recognition models,
with respect to their retention of information not relevant
for identification.  For example, the embeddings
produced by body models include information about
gender and camera angle, which can be linearly decoded from the network embeddings.  
Body networks also retain information
about the dataset from which an image originated. In principle, this is consistent
with the encoding of seemingly ``trivial'' image information, but it also opens the door to investigating exactly what information in the image can specify an originating dataset so accurately.   
We show that simple techniques previously used to boost identification performance are still applicable in the deep learning era. Finally, we provide an algorithm that substantially boosts our networks' retrieval performance using only a pre-defined gallery set.  
\subsection{Future Works}
The contribution of the face to body identification suggests opportunities for further improvement of face processing within
the context of a whole person image. This could be done
by training body networks to utilize face information more directly when it is available.
Further experimentation on the types of image features that support image-origin predictions could open new security applications, such as tracking image sources. Additionally, exploring the information embedded in body network representations may reveal potential security risks associated with body identification networks.
In the dimensionality reduction experiment, we focused on removing early principal components. Future research could be aimed at identifying optimal PC subspaces.

\section{ACKNOWLEDGMENTS}
This research is based upon work supported in part by
the Office of the Director of National Intelligence (ODNI),
Intelligence Advanced Research Projects Activity (IARPA),
via [2022-21102100005]. The views and conclusions contained herein are those of the authors and should not be
interpreted as necessarily representing the official policies,
either expressed or implied, of ODNI, IARPA, or the U.S.
Government. The US. Government is authorized to reproduce
and distribute reprints for governmental purposes notwithstanding any copyright annotation therein.


\section*{ETHICAL IMPACT STATEMENT}
We have read the guidelines for the Ethical Impact Statement.
The development of  body identification models does not 
 involve direct contact with human subjects, and therefore
 does not require approval by an Institutional Review Board. 
 Instead,  images/videos 
of human subjects are incorporated as training 
and test data for body identification models. We used only datasets 
(videos and images of people) that have been pre-screened and approved for ethical data collection standards by a United States government
funding agency, XXXX. The standards applied for dataset approval
require consent from the subjects who are depicted in the images/video for use in research.
Images/videos of subjects who
appear in publications require additional consent.
We followed these guidelines carefully. Images displayed in
the paper have been properly consented  and are displayed
according to the published  instructions for use of the dataset.

The development and study of biometric identification algorithms entails risk to individuals and societies. It is clear that these systems can have negative impacts if they are
misused. They can potentially threaten
individual  privacy and can impinge on  freedom of movement and expression in a society. 
The goal of our work is to better understand 
how these systems function. The results of this work can have both positive and negative societal 
impacts. On the positive side, knowing the types of representations created by body identification networks can help to minimize person identification errors. It can also serve to set reasonable performance expectations, thereby limiting the scope of use. On the negative side,  the knowledge gained can potentially be used to manipulate a 
system in unethical ways and to create synthetic images that can be misused or misinterpreted.

These risks are mitigated by the potential for positive societal impact. Body identification algorithms can  be used to locate missing people (including children). They can also be used in law enforcement to identify individuals implicated in crimes. Legitimate and societally-approved use can protect the general public from harm.  Of note, body identification systems can be used in combination with face identification systems to improve identification accuracy, thereby minimizing erroneous identifications.

{\small
\bibliographystyle{ieee}

}

\end{document}